\documentclass[twocolumn,9pt]{article} 

\usepackage[square,numbers,sort&compress,comma]{natbib}

\usepackage{amsmath}
\usepackage{amsfonts}
\usepackage{amssymb}
\usepackage{caption}
\usepackage{graphicx}
\usepackage{latexsym}
\usepackage{times}
\usepackage[pagewise]{lineno}
\usepackage{hyperref}
\usepackage{xcolor}
\usepackage[version=4]{mhchem}

\topmargin - 12pt 
\renewenvironment{abstract}%
              {
               \small
               {\bfseries \abstractname}
               \par
               \vspace{10pt}
              }

\renewcommand\abstractname{Abstract}

\newcommand{\nomenclature}
              [1]
              {
               \bgroup
               \flushleft
               \small\bf
               #1
               \par
               \egroup
              }

\renewcommand{\section}
              [1]
              {
               \bgroup
               \flushleft
               \small\bf
               \refstepcounter{section}
               \arabic{section}. #1
               \par
               \egroup
              }

\renewcommand{\subsection}
              [1]
              {
               \bgroup
               \flushleft
               \small\em
               \refstepcounter{subsection}
               \arabic{section}.
               \arabic{subsection}. #1
               \par
               \egroup
              }

\renewcommand{\subsubsection}
              [1]
              {
               \bgroup
               \flushleft
               \small\em
               \refstepcounter{subsubsection}
               \arabic{section}.
               \arabic{subsection}.
               \arabic{subsubsection}. #1
               \par
               \egroup
              }

  \newcommand{\acknowledgement}
              [1]
              {
               \bgroup
               \flushleft
               \small\bf
               #1
               \par
               \egroup
              }

  \newcommand{\sectionbib}
              [1]
              {
               \bgroup
               \flushleft
               \small\bf
               #1
               \par
               \egroup
              }

\begin{document}



\small
\baselineskip 10pt

\setcounter{page}{1}
\title{\LARGE \bf A graph neural network based chemical mechanism reduction method for combustion applications}

\author{{\large Manuru Nithin Padiyar$^{a}$, Priyabrat Dash$^{a}$, Konduri Aditya$^{a,*}$}\\[10pt]
        {\footnotesize \em $^a$Department of Computational and Data Sciences, Indian Institute of Science, Bengaluru, 560012, KA, India}\\[-5pt]}

\date{}  

\twocolumn[\begin{@twocolumnfalse}
\maketitle
\rule{\textwidth}{0.5pt}
\vspace{-5pt}

\begin{abstract}
Direct numerical simulations of turbulent reacting flows involving millions of grid points and detailed chemical mechanisms with hundreds of species and thousands of reactions are computationally prohibitive. To address this challenge, we present two data-driven chemical mechanism reduction formulations based on graph neural networks (GNNs) with message-passing transformer layers that learn nonlinear dependencies among species and reactions. The first formulation, GNN-SM, employs a pre-trained surrogate model to guide reduction across a broad range of reactor conditions. The second formulation, GNN-AE, uses an autoencoder formulation to obtain highly compact mechanisms that remain accurate within the thermochemical regimes used during training. The approaches are demonstrated on detailed mechanisms for methane (53 species, 325 reactions), ethylene (96 species, 1054 reactions), and iso-octane (1034 species, 8453 reactions). GNN-SM achieves reductions comparable to the established graph-based method DRGEP while maintaining accuracy across a wide range of thermochemical states. In contrast, GNN-AE achieves up to 95\% reduction in species and reactions and outperforms DRGEP within its target conditions. Overall, the proposed framework provides an automated, machine-learning-based pathway for chemical mechanism reduction that can complement traditional expert-guided analytical approaches.
\end{abstract}

\vspace{10pt}

{\bf Novelty and significance statement}

\vspace{10pt} 
This study introduces a graph neural network (GNN)-based framework for chemical mechanism reduction that combines graph-structured chemical representations with a pre-trained surrogate model and an autoencoder framework separately. The GNN learns nonlinear dependencies among species and reactions through message passing, identifying kinetically dominant pathways. One implementation utilizes a surrogate model to approximate combustion characteristics across a range of thermochemical states, while the other focuses on targeted, compact reduction at a single state, through the autoencoder. Applied to methane, ethylene and iso-octane, the frameworks achieve a significant decrease in species and reaction counts (up to 95\%), while preserving key reactor metrics. Through the two approaches, this work demonstrates adaptive mechanism reduction, balancing robustness and compactness alongside accuracy, facilitating integration with multi-fuel kinetics, turbulent combustion modeling, and data-driven mechanism refinement.

\vspace{5pt}
\parbox{1.0\textwidth}{\footnotesize {\em Keywords:} Graph neural networks; Mechanism reduction; Data-driven modeling}
\rule{\textwidth}{0.5pt}
*Corresponding author.
\vspace{5pt}
\end{@twocolumnfalse}]

\section{Introduction\label{sec:introduction}} \addvspace{10pt}

High-fidelity combustion simulations such as direct numerical simulations (DNS) and large-eddy simulations (LES) are instrumental in elucidating flow-chemistry interactions that inform turbulent combustion modeling and guide the design of efficient, low-emission engines. These simulations typically involve meshes of the order of millions to billions of grid points, with computational cost scaling exponentially with grid points. At each grid point, several conservation equations are solved, the complexity of which depends on the size of the chemical mechanism. Current high-fidelity combustion simulations employ mechanisms comprising tens of species and hundreds of reactions to achieve accurate results necessary to understand the combustion dynamics \cite{rieth_direct_2024,niemietz_direct_2023,dash_analysis_nodate}. However, beyond this scale, the computational cost becomes prohibitive even on state-of-the-art high performance computing platforms for practically relevant geometries.
For complex hydrocarbon fuels such as kerosene or gasoline, detailed chemical mechanisms contain hundreds of species and thousands of reactions \cite{narayanaswamy_component_2016, curran_comprehensive_2002}. These mechanisms are formulated to remain valid across a wide range of temperatures, pressures, and equivalence ratios. Consequently, mechanism reduction techniques are routinely employed to derive compact, condition-specific mechanisms suitable for high-fidelity combustion simulations \cite{lu_directed_2005,pepiot-desjardins_efficient_2008,jaravel_error-controlled_2019,sun_path_2010}.

Among the various approaches developed for mechanism reduction, graph-based methods have proven particularly effective due to their ability to capture the influence of neighboring species through network connectivity. Prominent examples include the directed relation graph (DRG) method \cite{lu_directed_2005} and its extensions, namely the directed relation graph with error propagation (DRGEP) \cite{pepiot-desjardins_efficient_2008} and the DRGEP with sensitivity analysis (DRGEP-SA) \cite{jaravel_error-controlled_2019} methods. While systematic, DRGEP-SA requires repeated sensitivity analyses over numerous steps. Neural network frameworks circumvent this requirement through GPU acceleration. Furthermore, traditional graph-based methods are often restricted to local, nearest-neighbor interactions and may therefore fail to capture long-range or global correlations in the system.

Graph neural networks (GNNs) offer a natural extension of these approaches by enabling the learning of complex, nonlinear relationships between species and reactions directly from data. Unlike traditional artificial neural networks (ANNs), which operate on fixed-size vector or matrix inputs, GNNs process data represented as graphs. In the context of chemical reaction mechanisms, nodes correspond to chemical species while edges represent their interactions through elementary reactions, allowing the model to directly encode the connectivity and relational structure of reaction networks.

Within this GNN framework, training relies on the key concept of message passing, whereby each node iteratively updates its latent representation by aggregating information from neighboring nodes and the connecting edges \cite{scarselli_graph_2009}. Through successive message-passing steps, information propagates across the graph, enabling the model to capture both local reaction dependencies and broader network-level correlations. Consequently, GNNs provide a powerful data-driven framework for analyzing and reducing chemical mechanisms, as well as for modeling complex chemical kinetics. Beyond combustion chemistry, GNNs have been successfully applied to atomistic-scale problems such as molecular dynamics \cite{li_graph_2022}, and to continuum-scale systems, including fluid flow modeling and super-resolution of turbulent fields \cite{dash_super-resolution_2026}.

In this work, we propose a GNN-based methodology for chemical mechanism reduction that combines the structural advantages of graph representations, as employed in DRG and DRGEP methodologies \cite{borde_mathematical_2025}, with the nonlinear function approximation capability of neural networks \cite{sen_large_2010}. Within this formulation, the chemical mechanism is represented as a graph in which species and reactions define the connectivity of the network, while the GNN learns the underlying dependencies governing species interactions directly from data. The resulting model identifies influential pathways and species contributions through data-driven message passing across the reaction network, enabling the systematic reduction of complex kinetic mechanisms into compact forms suitable for high-fidelity simulations. The objective is to obtain reduced mechanisms that retain the predictive fidelity of the detailed mechanism across relevant thermochemical conditions while significantly lowering the computational cost of reacting flow simulations.

\section{Methodology\label{sec:methodology}} \addvspace{10pt}
In this section, we describe the methodology underlying the proposed graph neural network (GNN)-based {chemical} mechanism reduction strategy, which is implemented through two distinct formulations. The first approach, denoted as GNN-SM, leverages the evolution of thermochemical states across multiple reactor configurations using a pre-trained surrogate model. The second approach, GNN-AE, employs an autoencoder-based architecture trained on discrete thermochemical states to learn a relatively more compact representation of the reacting system. The theoretical basis of each formulation is described below, followed by the specific network architectures and the corresponding training procedures.

\subsection{GNN reduction methodology}\label{reduction workflow}\addvspace{10pt}

\begin{figure*}
    \centering
    \includegraphics[width=400pt]{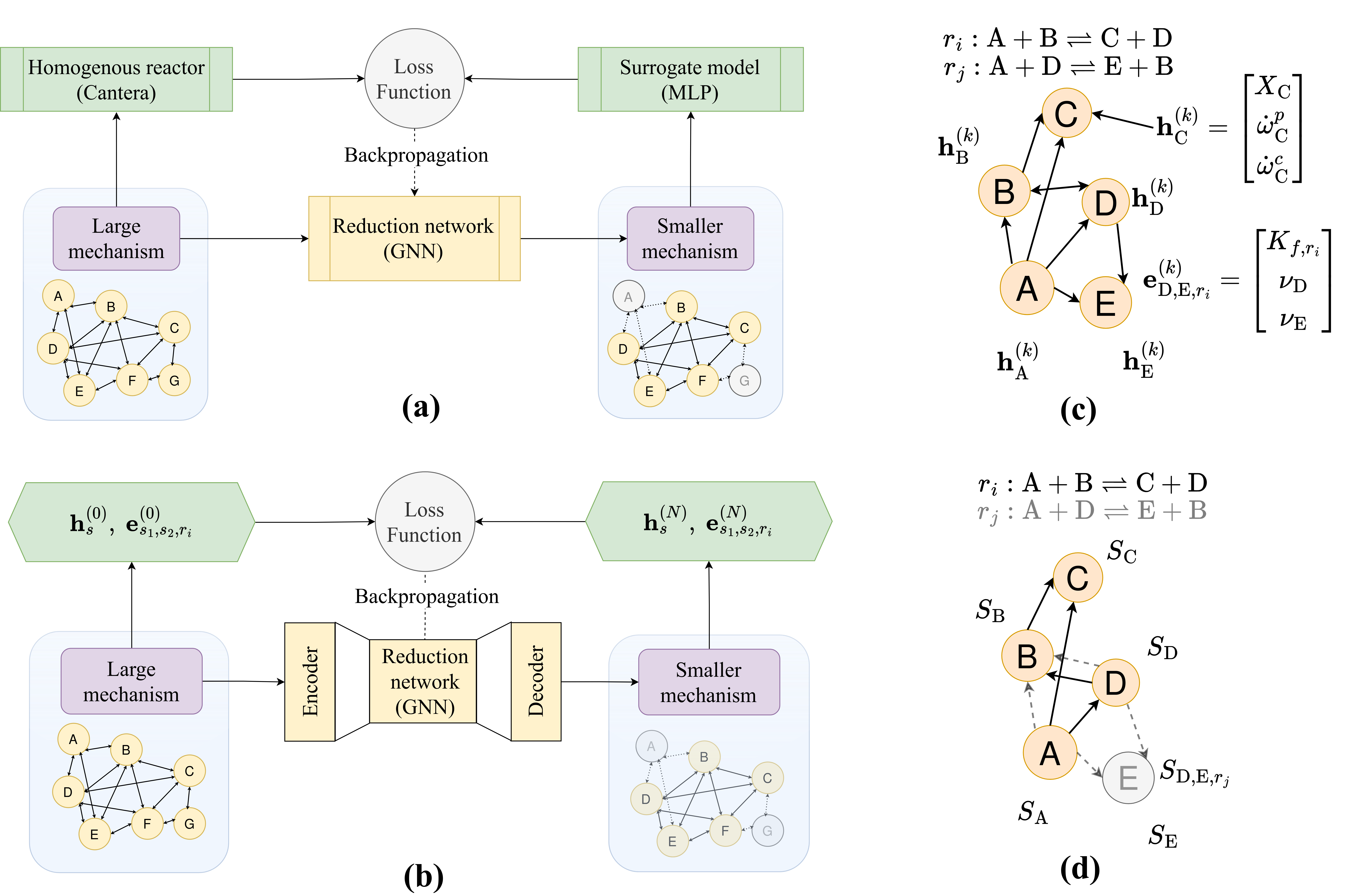}
    \caption{\footnotesize A flowchart representation of the (a) GNN-SM implementation, (b) GNN-AE implementation along with graphs of the (c) input and (d) output mechanisms.}
    \label{fig:flowchart}
\end{figure*}

The proposed GNNs for mechanism reduction operate on a graph representation of the chemical kinetics. The input graph contains node and edge feature vectors constructed from quantities such as stoichiometric coefficients and instantaneous thermochemical scalars, including species concentrations, net reaction rates, and species production rates. Through message passing across the reaction network, the GNN processes these features and produces an output graph with identical connectivity but augmented with scalar importance scores associated with each node and edge. These scores are subsequently used to identify the species and reactions that exert the strongest influence on the system dynamics, thereby guiding the reduction of the detailed mechanism. A schematic overview of the proposed GNN-based reduction methodology is shown in Fig.~\ref{fig:flowchart}.

The chemical mechanism is represented as a directed graph in which nodes correspond to chemical species and edges denote interactions between species participating in the same reaction, all at a chosen state, as illustrated in Fig.~\ref{fig:flowchart}(c). Directed edges encode the influence of one species on the formation or consumption of another, thereby preserving the structure of reaction pathways and the propagation of chemical influence through the network. Each species node $\mathrm{X}$ is associated with a feature vector $\mathbf{h}_{\mathrm{X}}^{(k)}$ comprising the species mole fraction at the current thermochemical state and its rates of production and consumption\footnote{The superscript $(k)$ denotes the representation at the $k^\text{th}$ GNN layer.}. Edge feature vectors $\mathbf{e}_{\mathrm{X,Y},r_i}^{(k)}$ encode reaction-level information such as stoichiometric coefficients and kinetic parameters for reaction $r_i$, thereby incorporating both the topological and kinetic structure of the mechanism without directly encoding the thermochemical state into the network.

The GNN produces an output graph with the same node–edge connectivity, but with scalar importance scores $S_\mathrm{X}$ for species and $S_{\mathrm{X,Y},r_i}$ for reactions, as illustrated in Fig.~\ref{fig:flowchart}(d). These scores, bounded between 0 and 1, quantify the relative importance of species and reactions and form the basis for mechanism reduction. Species and reactions with low importance scores can be pruned either by retaining a specified fraction of species (or reactions) or by applying a threshold to the predicted scores. The GNN-SM formulation adopts the former strategy, whereas GNN-AE employs the latter. The reduction procedure is performed iteratively to balance mechanism compactness with predictive accuracy, ensuring that the dominant reaction pathways governing the target combustion behavior are preserved. The two formulations also differ in the type of training data employed and in the evaluation of the loss function. The specific details of each approach are described next.

\noindent \textbf{GNN-SM}:
To construct reduced mechanisms valid for target applications, the GNN must be trained using representative thermochemical states from canonical configurations such as well-stirred reactors or counterflow flames. However, standard solvers for these systems are non-differentiable, preventing direct back-propagation during training. The GNN-SM framework circumvents this limitation through a two-stage approach. First, a multi-layer perceptron (MLP) surrogate model is trained offline using the comprehensive mechanism. The surrogate network learns to map initial thermochemical states to target quantities of interest, such as ignition delay time and laminar flame speed. By doing so, the surrogate model provides a differentiable pathway for loss evaluation. To ensure robustness, the surrogate training dataset is generated from canonical simulations spanning a broad envelope of initial temperatures, pressures, and equivalence ratios.

In the second stage, the GNN is optimized to extract reduced mechanisms tailored to specific target conditions provided during training. To represent the flame, the input graph is initialized at a given thermochemical state corresponding to a particular region within the flame structure. During training, these initialization points are sampled across multiple flame regions, as well as across different flame configurations, to ensure coverage of key regimes. The state‑dependent quantities are mapped into the node and edge features of the graph provided to the GNN. The network processes these features and outputs scalar importance scores for each species node and reaction edge. A top‑$k$ pooling operation then selects the most influential species and their associated reactions based on these scores, yielding a reduced representation of the mechanism graph that preserves the dominant reaction pathways at the target conditions.

Subsequently, a global pooling operation is used to estimate the initial thermochemical conditions, specifically the initial temperature $\hat{T}$ and pressure $\hat{P}$, from the node and edge features of the reduced mechanism graph. This effectively trains the GNN to infer the unburnt state or reactor initial condition corresponding to a given flame region, that is, to backtrack from an intermediate thermochemical state to its initial $T,P$ values. The pooled quantities are then passed to a pre-trained surrogate model, which predicts the target combustion quantity, such as the ignition delay time $\hat{\tau}_{ig}$. The corresponding reference value $\tau_{ig}$ is computed \textit{a priori} using the detailed mechanism. The prediction error is quantified via an absolute error: $\mathcal{L}_{ig} = \left| \hat{\tau}_{ig} - \tau_{ig} \right|$. To promote sparsity and minimize the size of the retained mechanism, an additional regularization term based on the predicted importance scores is introduced. The sparsity loss is defined as the mean of the scalar importance scores across all $N_{\text{nodes}}$ species nodes and $N_{\text{edges}}$ reaction edges in the complete graph:
\begin{equation}\label{sparsity}
\mathcal{L}_{\text{sparsity}} 
= \frac{1}{N_{\text{nodes}}} \sum_{\mathrm{X}} S_{\mathrm{X}} 
+ \frac{1}{N_{\text{edges}}} \sum_{\mathrm{X,Y},r_i} S_{\mathrm{X,Y},r_i}.
\end{equation}
To prevent excessive pruning that would degrade predictive accuracy, this term is weighted by a scaling factor $\lambda_{\text{sparsity}}$. The total loss function governing the GNN optimization is therefore $ \mathcal{L}_{\text{total}} = \mathcal{L}_{ig} + \lambda_{\text{sparsity}} \mathcal{L}_{\text{sparsity}}$.
\addvspace{10pt}

\noindent \textbf{GNN-AE}: While the GNN-SM framework leverages the evolution of thermochemical states and a surrogate model to ensure broad validity across operating envelopes, the autoencoder-based \cite{kipf2016variationalgraphautoencoders} formulation (GNN-AE) takes a more localized, purely data-driven approach. By focusing on the reconstruction of discrete thermochemical states, the GNN-AE framework is expected to achieve significantly more compact reduced mechanisms relative to the GNN-SM strategy. This high degree of mechanism compression is enabled by training a graph autoencoder to learn an identity mapping of the localized thermochemical state. The input node and edge features remain identical to those in the GNN-SM framework. During the encoding phase, the input graph representing the comprehensive kinetic mechanism is processed through message passing layers and projected into a lower-dimensional latent space. Crucially, the architecture is structured such that this latent representation directly corresponds to the scalar importance scores ($S_\mathrm{X}$ for species and $S_{\mathrm{X,Y},r_i}$ for reactions). The decoder subsequently relies on these latent importance scores to reconstruct the original, unreduced instantaneous thermochemical quantities such as species mole fractions and net reaction rates. The primary objective is to minimize the reconstruction error of the node and edge features, defined by their respective $L_2$ norms:
\begin{equation}
\mathcal{L}_{n} = |\mathbf{h}_{\mathrm{X}}^{(N)} - \mathbf{h}_{\mathrm{X}}^{(0)}|_2
\end{equation}
\begin{equation}
\mathcal{L}_{e} = |\mathbf{e}_{\mathrm{X,Y},r_i}^{(N)} - \mathbf{e}_{\mathrm{X,Y},r_i}^{(0)}|_2,
\end{equation}
where the superscripts $(0)$ and $(N)$ denote the features at the input layer and the final output layer, respectively. Analogous to the GNN-SM formulation, a sparsity constraint is imposed to penalize non-essential pathways and drive the reduction process. By minimizing the reconstruction loss alongside this sparsity regularization term, the network is compelled to assign high importance scores only to the species and reactions strictly required to preserve the local chemical dynamics. This sparsity loss explicitly evaluates the mean scores across all species and reactions and is computed as in Eq.~\ref{sparsity}. The individual loss components are aggregated using specified weighting hyperparameters to form the total loss:
\begin{equation}
    \mathcal{L}_{\text{total}} = \lambda_{\text{reconstruction}}(\mathcal{L}_n + \mathcal{L}_e) + \lambda_{\text{sparsity}}\mathcal{L}_{\text{sparsity}}.
\end{equation}
Upon optimization, the framework yields an ordered list of importance scores for all species nodes. The reduced mechanism is extracted by applying a predefined threshold to these latent scores, dynamically pruning any nodes and edges that fall below the cutoff. To ensure physical robustness, this pruning is conducted iteratively. 

To achieve further compaction of the skeletal mechanism, a greedy search algorithm is employed as a secondary reduction step. This procedure iteratively evaluates the removal of each remaining species by monitoring the resulting deviations in critical combustion metrics, specifically the ignition delay time and the peak heat release rate. If the elimination of a candidate species induces an error that exceeds a prescribed tolerance for either metric, the species is classified as essential to the active chemical pathways and is selectively retained. 

\subsection{Network architectures}\label{message-passing}\addvspace{10pt}

\noindent \textbf{GNN-SM Architecture}: The GNN employed in this formulation consists of two graph transformer layers each with a hidden dimension of 128. The resulting graph embeddings are routed into three distinct task-specific heads. The first two heads are multi-layer perceptrons (MLPs) responsible for predicting the node and edge scalar importance scores, respectively. Based on a specified retention ratio, a top-$k$ selection isolates the most critical species and reactions to construct a sparse subgraph. This pruned subgraph is subsequently processed by the third head, which performs a global pooling operation to estimate the initial unburnt thermochemical conditions, $\hat{T}$ and $\hat{P}$. Simultaneously, the continuous importance scores are used to evaluate the sparsity penalty. The overall GNN is trained for a prescribed number of epochs ($N_{\text{epochs}}$) using the Adam optimizer, during which the weights of the pre-trained surrogate model remain strictly frozen. By replacing the ignition delay loss with a performance metric representative of a different configuration, the methodology can be directly applied to other canonical flame configurations or reactors.\\

\noindent \textbf{GNN-AE Architecture}: The GNN-AE formulation utilizes a purely data-driven, message-passing encoder-decoder architecture. The initial node and edge feature vectors, possessing dimension $N_{\text{features}}$, are first embedded into a 128-dimensional hidden space. Message passing is then executed across the graph topology to capture the highly nonlinear thermochemical interactions among participating species. Subsequently, these hidden representations are projected onto a one-dimensional latent space, explicitly yielding the node and edge importance scores. These scores are extracted as the primary output for mechanism reduction while simultaneously being fed into the decoder. To evaluate the reconstruction loss, the decoder expands these latent scores through another 128-dimensional MLP to reconstruct the original $N_{\text{features}}$-dimensional thermochemical state vectors. The network weights are optimized over a prescribed number of epochs ($N_{\text{epochs}}$) using the Adam optimizer.\\

\noindent\textbf{Elements of GNN}: The key feature of GNNs is pairwise message passing, wherein the node and edge features are iteratively updated based on the state of the neighboring nodes and edges. The exact computation of the updated iteration varies depending upon the GNN layer chosen \cite{kipf_semi-supervised_2017,velickovic_graph_2018,shi_masked_2021}. In our work, we utilize graph transformer layers for their attention capabilities as well as their ability to incorporate edge features in message passing.

For a node in a graph $u$ with the corresponding node features $\textbf{h}_u$, the message passing step of proceeds as
$
    \textbf{h}_u' = \textbf{W}_1\textbf{h}_u + \sum_{v\in \mathcal{N}(u)}\alpha_{u,v}\textbf{W}_2\textbf{h}_v,
$
where $\mathcal{N}(u)$ denotes the set of all nodes that neighbor the node $u$. The attention coefficients $\alpha_{u,v}$ are computed as:
\begin{equation}
    \alpha_{u,v} = \text{softmax}\left(\frac{(\textbf{W}_3\textbf{h}_u)^\top(\textbf{W}_4\textbf{h}_v)}{\sqrt{d}}\right).
\end{equation}
Since we seek to incorporate edge features into the computation, the above equation takes the form:
\begin{equation}
    \alpha_{u,v} = \text{softmax}\left(\frac{(\textbf{W}_3\textbf{h}_u)^\top(\textbf{W}_4\textbf{h}_v+\textbf{W}_5\textbf{e}_{u,v})}{\sqrt{d}}\right).
\end{equation}
In the above equations, the various $\textbf{W}_i$'s indicate the matrices of trainable parameters.

\section{Results\label{sec:results}} \addvspace{10pt}

To assess the performance of the proposed framework, we consider the reduction of three comprehensive chemical mechanisms. These mechanisms are also reduced using the standard DRGEP method as a baseline. For each mechanism, five variants are evaluated: the original comprehensive mechanism, the DRGEP-reduced mechanism, the most compact mechanism obtained using GNN-SM, and the mechanisms produced by GNN-AE before and after greedy search refinement.

\subsection{Methane oxidation}\addvspace{10pt}

First, reduced mechanisms for methane combustion are developed using the comprehensive GRI-Mech 3.0 mechanism \cite{smith_gri-mech_nodate}, which contains 53 species and 325 reactions. The baseline DRGEP method yields a reduced mechanism consisting of 23 species and 123 reactions. 
For the GNN-SM approach, training datasets are generated over a temperature range of $1300$--$1700$~K at $1$~atm and stoichiometric equivalence ratio. In contrast, the GNN-AE approach is trained using thermochemical states sampled around $1500$~K at the same pressure and equivalence ratio. The GNN-SM approach employs an iterative reduction procedure in which Pass~1 and Pass~2 produce mechanisms containing 35 and 29 species, respectively, while Pass~3 further reduces the mechanism to 27 species with a marginal ignition delay time (IDT) error of 1.45\%. The GNN-AE approach initially produces a 35-species mechanism with an IDT error of 1.77\%, which is subsequently refined to a highly compact 19-species mechanism (GNN-AE-G) through targeted greedy search. Although GNN-AE-G achieves the most aggressive reduction, it incurs a larger average IDT error of 6.83\% over the temperature range 1000--2000~K at atmospheric pressure and stoichiometric conditions, as summarized in \autoref{mechanisms_comparison}. This temperature range captures both high-temperature chain-branching kinetics and low-to-intermediate temperature pathways involving the formation and decomposition of \ce{HO2}.

\begin{table}[h]
\footnotesize
\centering
\caption{Comparison of chemical mechanisms for methane oxidation (1000K to 2000K).}
\label{mechanisms_comparison}
\begin{tabular}{lccc}
\hline
\textbf{Mechanism} & \textbf{$\mathbf{n_s/n_r}$} & \textbf{Avg. IDT Error (\%)} \\ \hline
GRI-Mech 3.0   & 53 / 325  & --- \\
DRGEP      & 23 / 123  & 1.53 \\ \hline
\multicolumn{3}{l}{\textbf{GNN-SM}} \\ \hline
Pass 1      & 35 / 189  & --- \\
Pass 2      & 29 / 177  & --- \\
Pass 3          & 27 / 157  & 1.45 \\ \hline
\multicolumn{3}{l}{\textbf{GNN-AE}} \\ \hline
GNN-AE   & 35 / 189  & 1.77 \\
GNN-AE-G    & 19 / 82   & 6.83 \\ \hline
\end{tabular}
\end{table}

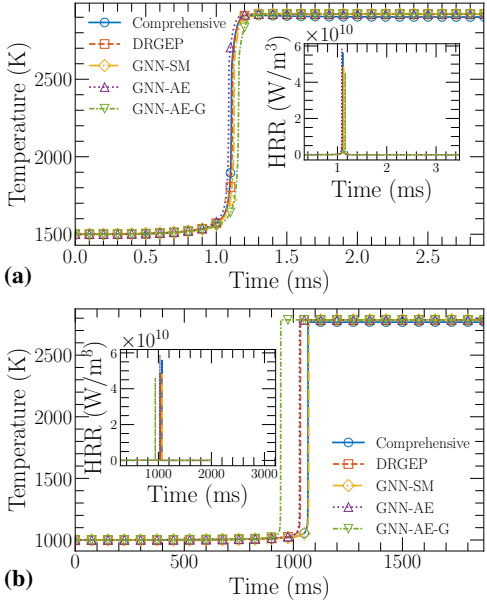
\begin{figure}[h!]
    \centering
    \includegraphics[width=185pt]{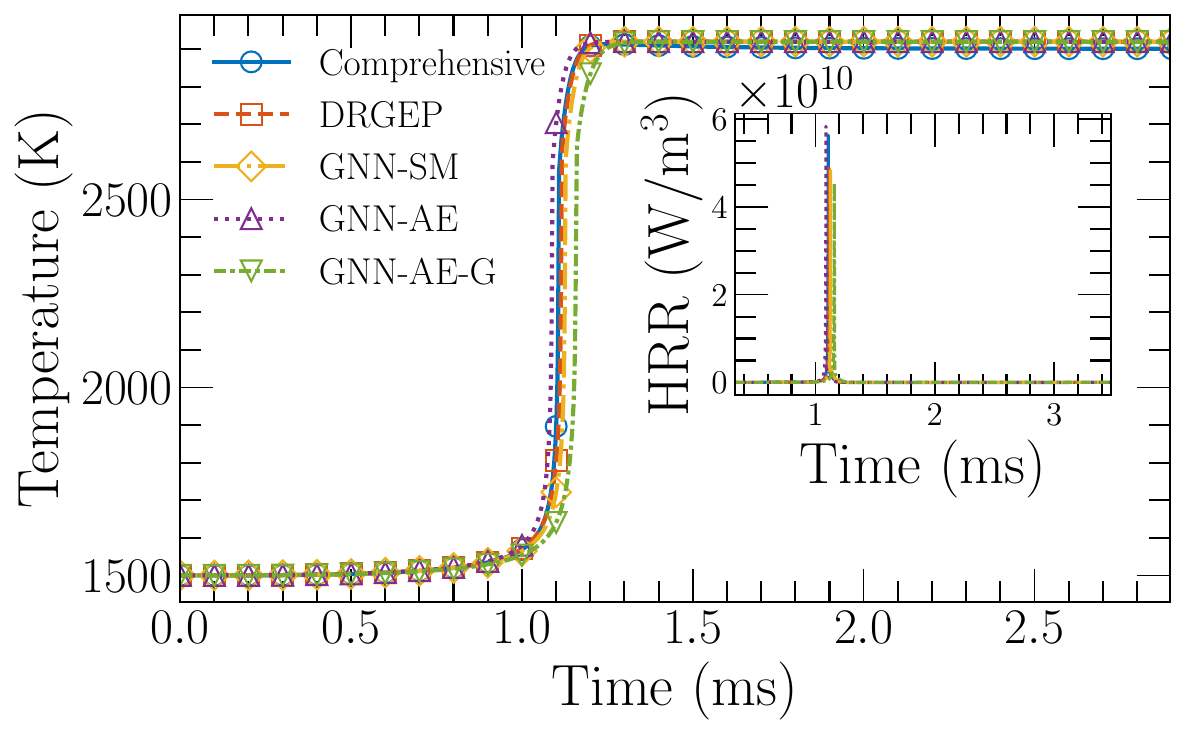}
    \includegraphics[width=185pt]{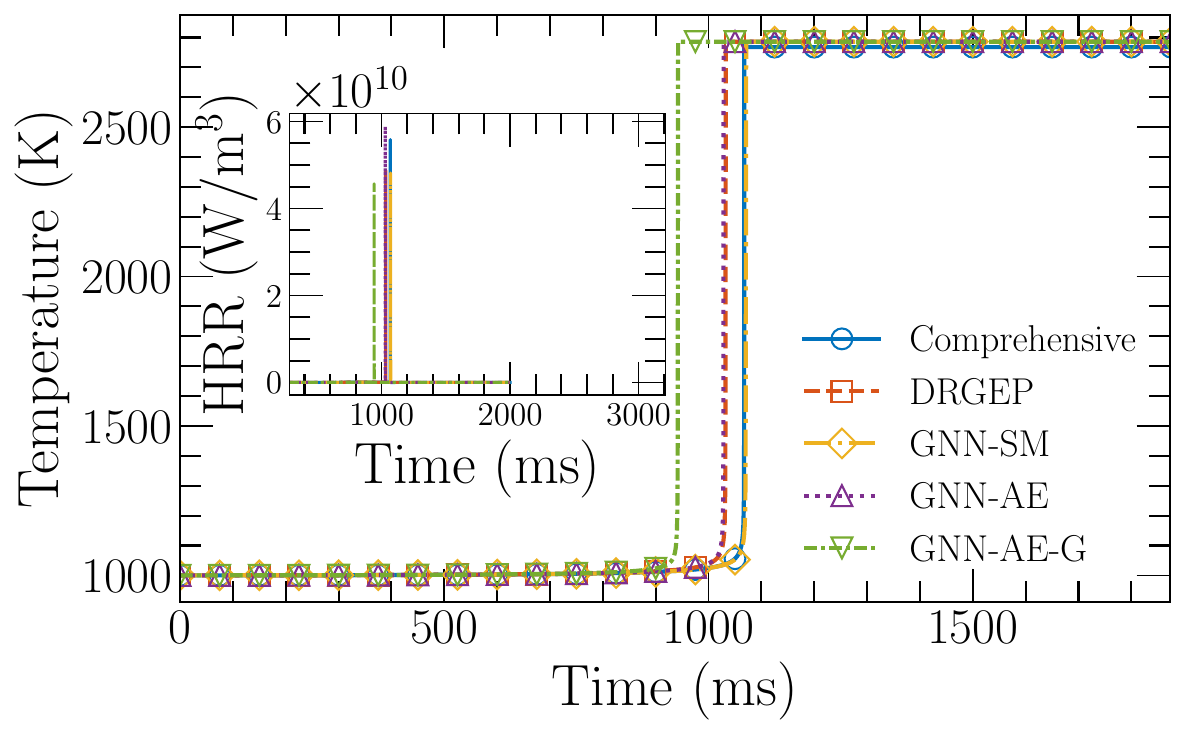}
    \caption{Temporal evolution of temperature and heat release rate at initial temperatures of (a) 1500K and (b) 1000K for methane oxidation.}
    \begin{picture}(0,0)
            \put(-90,160){\bf \small (a)}
            \put(-90,45){\bf \small (b)}
    \end{picture}
    \label{fig:methane-tempvstime}
\end{figure}

The temporal evolution of temperature and heat release rate (HRR) at an initial temperature 1500 K and 1000 K in an autoignition simulation, presented in Fig.~\ref{fig:methane-tempvstime}, illustrates the regime-dependent fidelity of the reduced mechanism. At 1500 K, all mechanisms, including the highly compact GNN-AE-G, exhibit excellent agreement with the comprehensive mechanism. In this high-temperature regime, kinetics are primarily driven by fast radical chain-branching steps, such as \ce{H + O2 <=> OH + O}, which are readily identified and preserved by all reduction methodologies. However, at 1000 K, a distinct deviation in performance is observed. While GNN-SM and DRGEP accurately capture the ignition delay and peak HRR, GNN-AE-G exhibits a significant error in ignition delay and a suppressed HRR profile.

This performance discrepancy is linked to the structure of the reduced mechanisms. Analysis of the GNN-AE-G mechanism reveals the aggressive pruning of \ce{C2} ``bridge'' intermediates, such as \ce{C2H2} and \ce{C2H3}, which are vital for radical propagation at intermediate temperatures. Conversely, the GNN-SM mechanism (Pass 3) retains a more comprehensive set of oxygenated species, including \ce{H2O2} and \ce{CH3OH}. At 1000 K, the accumulation and subsequent decomposition of \ce{H2O2} (\ce{H2O2 + M <=> OH + OH + M}) acts as the primary trigger for ignition. By capturing these slower initiating reactions, GNN-SM remains robust across a wider section of the state space, while GNN-AE-G provides a highly compact reduction strictly tailored to the high-temperature regime.

\begin{figure}
    \centering
    \includegraphics[width=185pt]{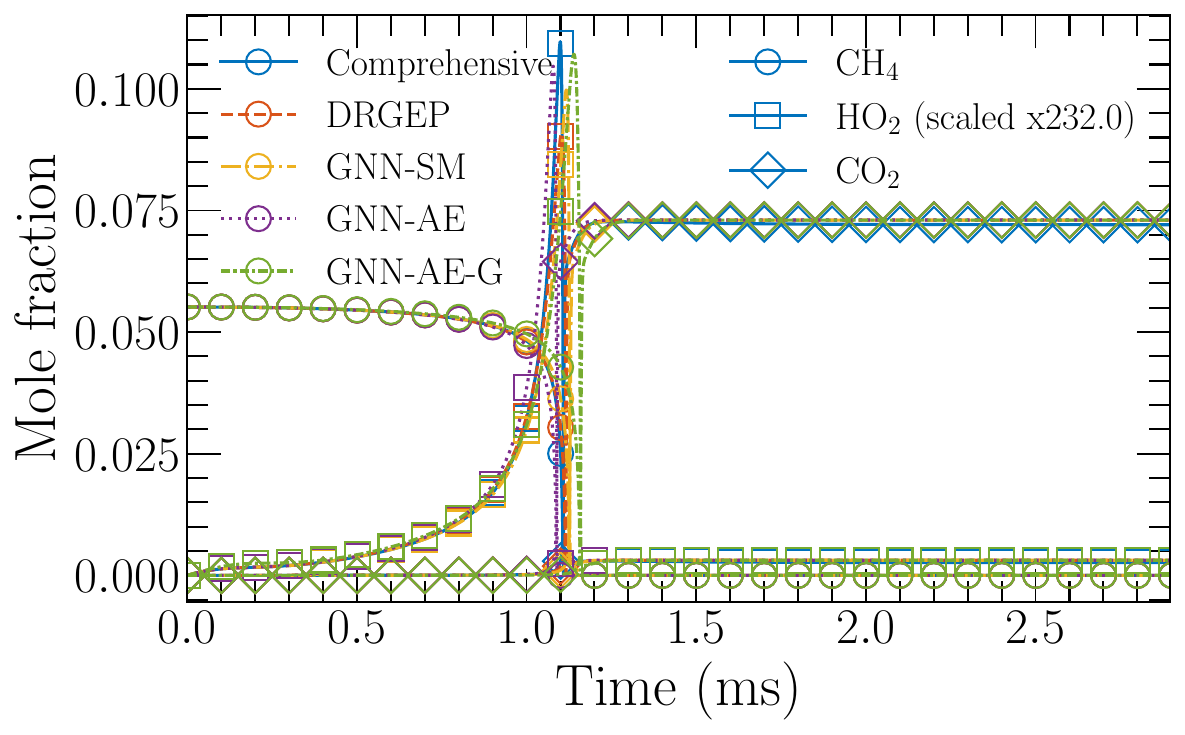}
    \caption{Temporal evolution of scaled species mole fractions at initial temperature of 1500K for methane oxidation.}
    \label{fig:methane X vs T}
\end{figure}

\begin{figure}
    \centering
    \includegraphics[width=185pt]{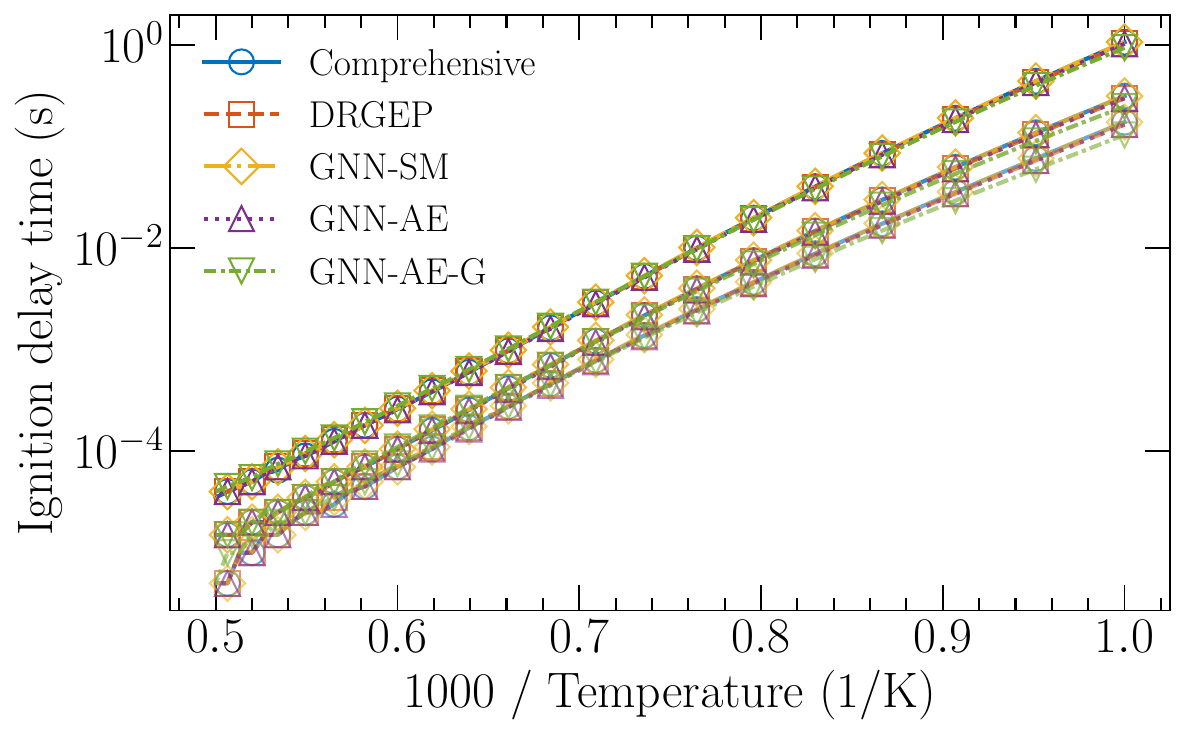}
    \caption{Validation of ignition delay times for the reduced methane mechanisms over a range of initial temperatures and pressures.}
    \label{fig:methane igt vs T}
\end{figure}

The fidelity of the radical pool is further validated in Fig.~\ref{fig:methane X vs T}, where the mass fractions of \ce{CH4}, \ce{CO2}, and \ce{HO2} are compared. The GNN-SM reduced mechanism shows near-perfect overlap with results obtained using the comprehensive mechanism, particularly in capturing the transient \ce{HO2} peak which is critical for intermediate-temperature reactivity. Finally, the global performance across a temperature range is shown in Fig.~\ref{fig:methane igt vs T}. Both GNN-SM and DRGEP conform with the baseline across the entire range, while GNN-AE and GNN-AE-G mechanisms demonstrate a sharp deviation at lower temperatures. This highlights the necessity of the surrogate model in GNN-SM to ensure kinetic robustness across diverse operating conditions.

\subsection{Ethylene oxidation} \addvspace{10pt}

The mechanism reduction capabilities of the GNN frameworks are further evaluated using a more complex FFCM-2 mechanism \cite{ZDV2023} (typically used for modeling ethylene oxidation), which serves as the comprehensive baseline with 96 species and 1054 reactions, with training data for both GNN-SM and GNN-AE being from the same range as was for methane.  Ethylene is a critical fuel for high-speed propulsion applications, requiring high-fidelity kinetics across large thermochemical gradients. As summarized in \autoref{ethylene mechanisms_comparison}, the DRGEP baseline reduces the system to 38 species and 288 reactions with an average ignition delay time (IDT) error of 11.56\%. In comparison, a six-pass GNN-SM reduction yields a more compact 36-species mechanism that achieves a lower IDT error of 8.15\%, demonstrating its superior ability to identify and prune the non-essential pathways in complex hydrocarbon systems. The autoencoder-based frameworks provide even more targeted reductions. While the general GNN-AE model (35 species) maintains an error of 8.99\%, the application of greedy search leads to highly specialized mechanisms. The GNN-AE-G (1500 K) model reaches a minimum of 25 species but suffers from a high average error of 19.71\% due to its poor generalizability. To address this, a second specialized mechanism, GNN-AE-G (1200 K), was developed, a 28-species model which yields a improved average error of 13.92\% by better capturing the intermediate-temperature regime.

\begin{table}[h!]
\footnotesize
\centering
\caption{Comparison of mechanisms for ethylene oxidation (1000K to 2000K).}
\label{ethylene mechanisms_comparison}
\begin{tabular}{lccc}
\hline
\textbf{Mechanism} & \textbf{$\mathbf{n_s/n_r}$} & \textbf{Avg. IDT Error (\%)} \\ \hline
FFCM2   & 96 / 1054  & --- \\
DRGEP       & 38 / 288  & 11.56 \\ \hline
\multicolumn{3}{l}{\textbf{GNN-SM}} \\ \hline
Pass 1      &  77/814   & --- \\
Pass 3      &  44/397   & --- \\
Pass 5      &  37/302   & --- \\
Pass 6          & 36 / 292  & 8.15 \\ \hline
\multicolumn{3}{l}{\textbf{GNN-AE}} \\ \hline
GNN-AE   & 35 / 288  & 8.99 \\
AE-G (1500K)    & 25 / 175   &  19.71\\
AE-G (1200K) & 28 / 181 &  13.92\\ \hline
\end{tabular}
\end{table}

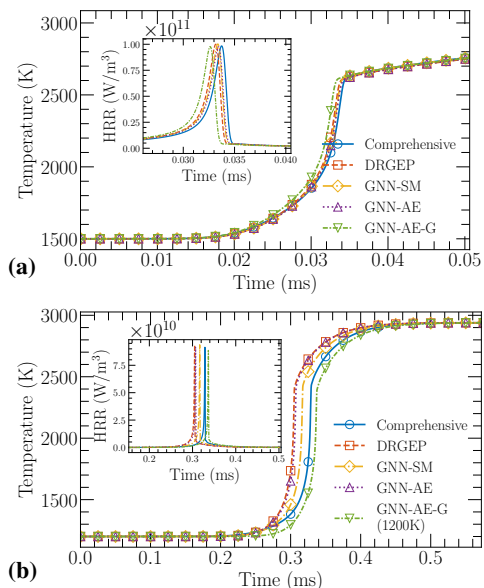
\begin{figure}[h!]
    \centering
    \includegraphics[width=180pt]{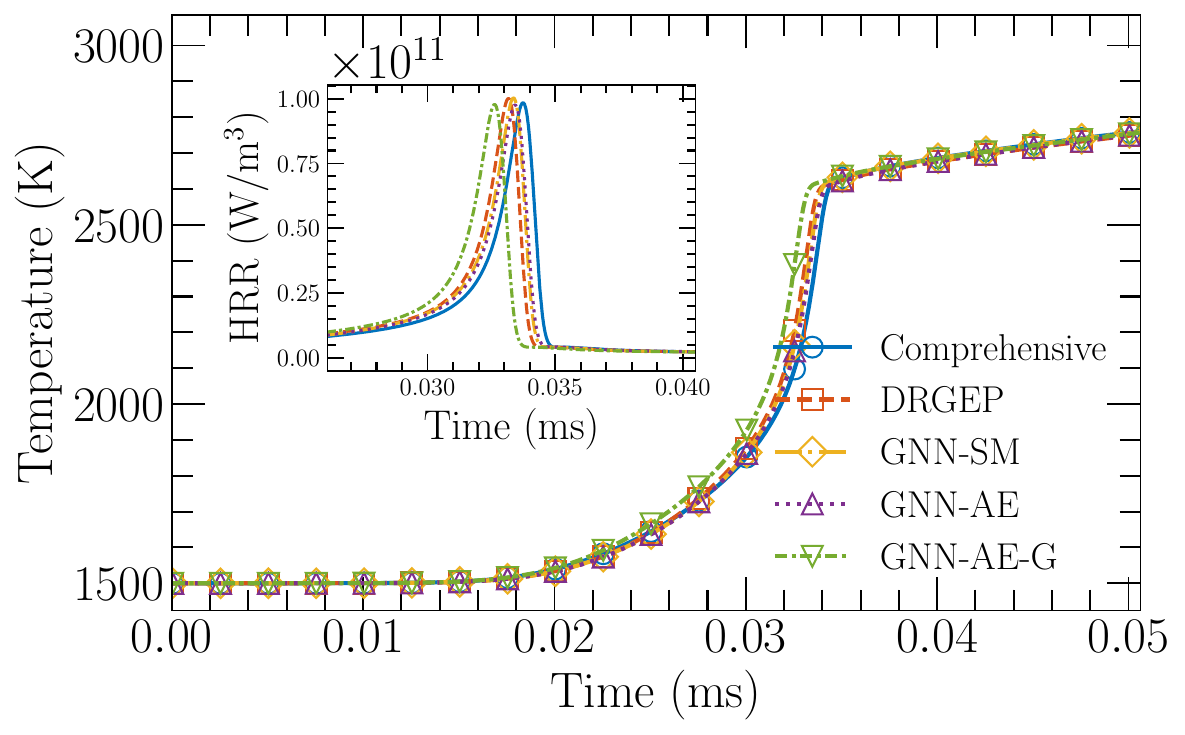}
    \includegraphics[width=180pt]{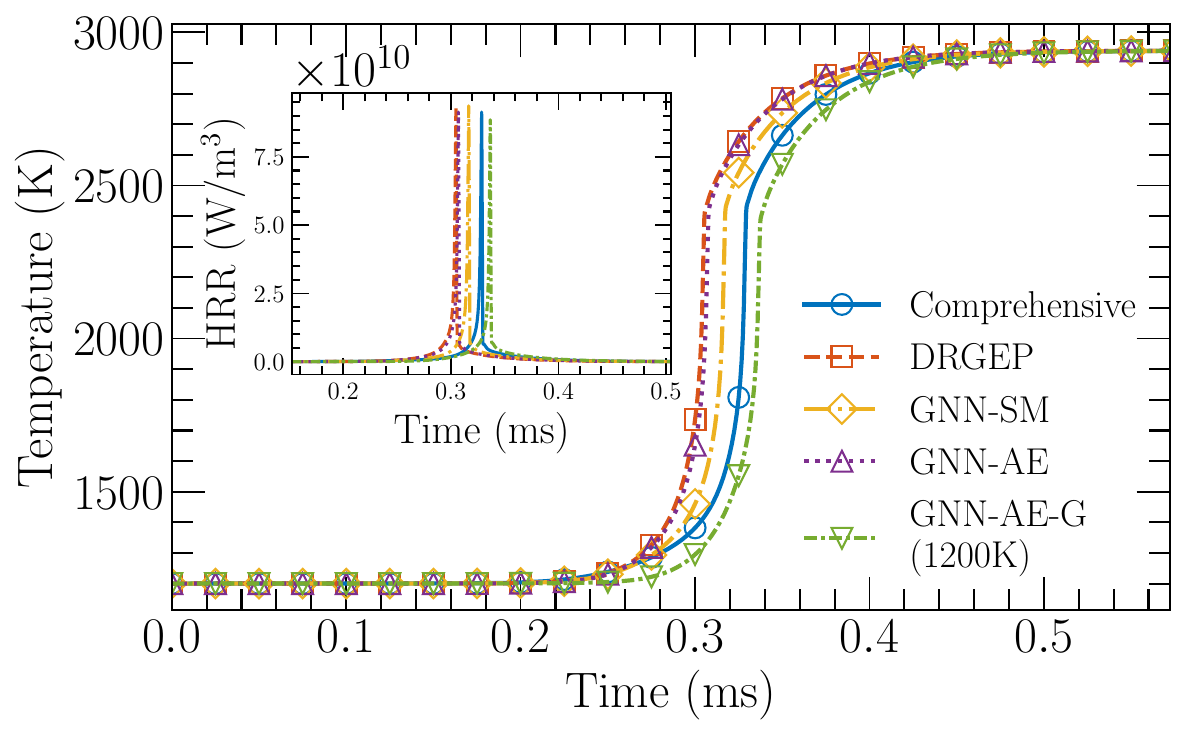}
    \caption{Temporal evolution of temperature and heat release rate at initial temperatures of (a) 1500K and (b) 1200K for ethylene oxidation.}
    \begin{picture}(0,0)
            \put(-90,160){\bf \small (a)}
            \put(-90,45){\bf \small (b)}
    \end{picture}
    \label{fig:ethylene temp vs time}
\end{figure}

The temporal evolution of temperature and heat release rate at 1500 K, shown in Fig.~\ref{fig:ethylene temp vs time}, highlights the trade-off between specialization and robustness. In these plots, the comprehensive (FFCM-2) baseline is compared against DRGEP, GNN-SM, GNN-AE, and the two specialized GNN-AE-G models. At 1500 K, the GNN-AE-G (1500 K) mechanism provides an exceptionally accurate reconstruction of the ignition event. However, as the temperature is lowered to 1200 K, this model exhibits a significant deviation, severely over-predicting the ignition delay. In contrast, the GNN-AE-G (1200 K) mechanism shows substantially better performance at this lower temperature, aligning closely with the comprehensive trend. This improvement stems from the model's training data, which forces the retention of 28 species compared to the 25 in the 1500 K model. Analysis of the reaction topologies reveals that the 1200 K model retains key intermediate pathways that the 1500 K model prunes as non-essential for high-temperature chemistry.

\begin{figure}[h!]
    \centering
    \includegraphics[width=180pt]{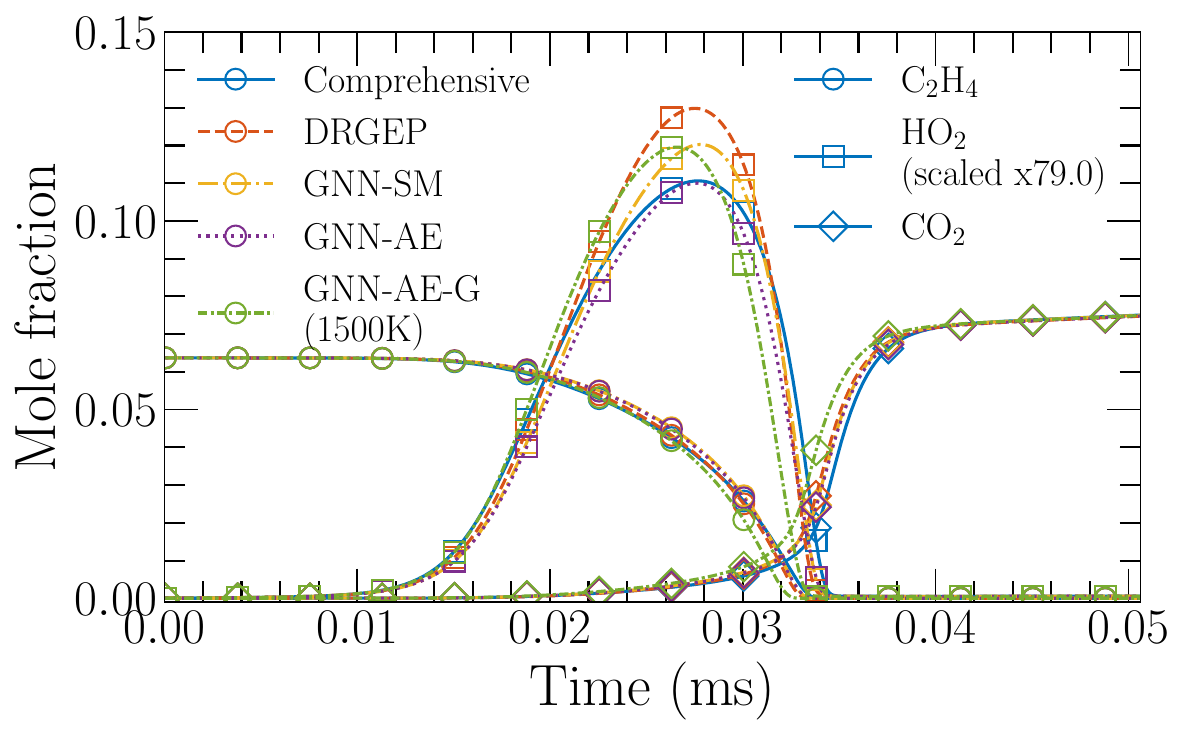}
    \caption{Temporal evolution of species mole fractions at initial temperature of 1500K for ethylene oxidation.}
    \label{fig:ethylene X}
\end{figure}

\begin{figure}[h!]
    \centering
    \includegraphics[width=180pt]{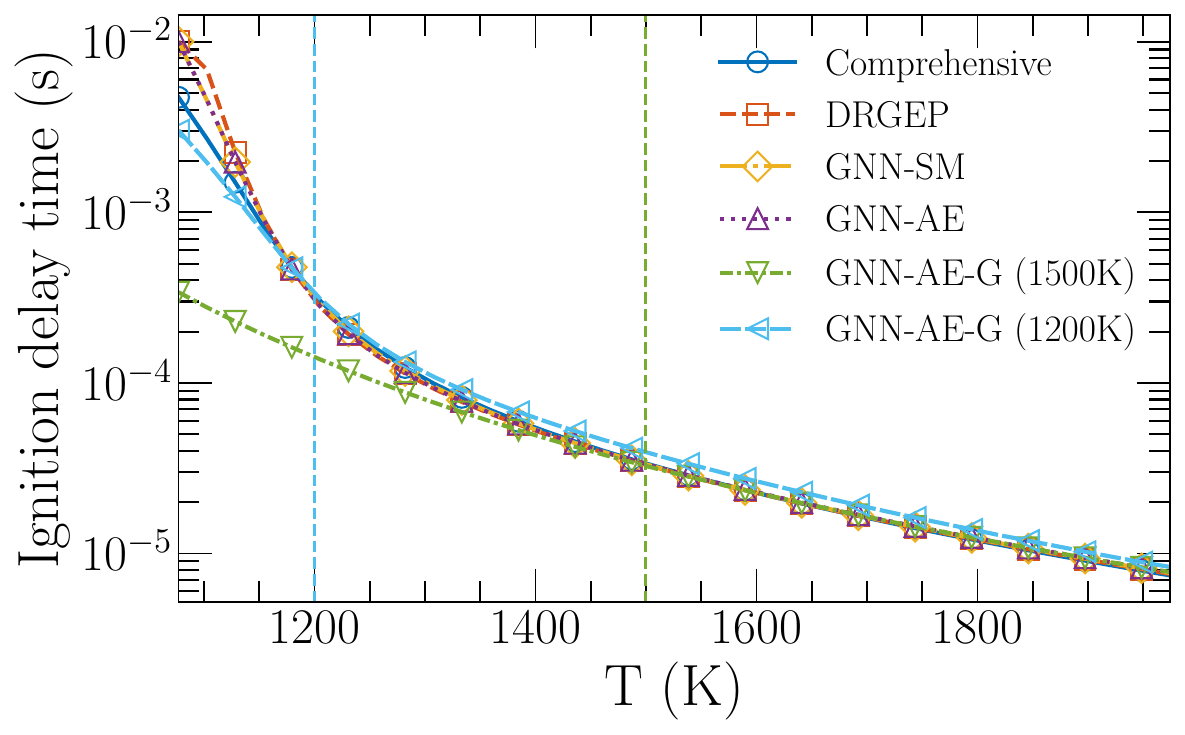}
    \caption{Validation of ignition delay times for reduced ethylene mechanisms across conditions. The dotted cyan and green lines indicate the regions around which the two GNN-AE-G mechanisms were reduced.}
    \label{fig:ethylene igt vs T}
\end{figure}

The underlying chemical reasoning for the success of GNN-SM as a more generalizable reduction strategy lies in its aggressive but strategic pruning. While GNN-SM identifies the \ce{C3} hydrocarbon chain (e.g., propane and propene pathways) as non-essential for ethylene target observables, it consistently preserves oxygenated intermediates such as \ce{H2O2} and \ce{CH3OH}. These species are vital for low-to-intermediate temperature reactivity, where the decomposition of hydrogen peroxide serves as the primary radical source. The exclusion of these pathways in the 1500 K specialized model explains its rapid performance degradation at 1200 K.

The species mole fraction profiles in Fig.~\ref{fig:ethylene X} further confirm this fidelity. Both the GNN-SM and the 1200 K-targeted GNN-AE-G model exhibit near-perfect tracking of \ce{C2H4} consumption and the evolution of \ce{CO2} and \ce{HO2} at the 1200 K condition. Finally, the global validation of ignition delay times across the temperature range in Fig.~\ref{fig:ethylene igt vs T} illustrates that while the GNN-AE-G models are superior within their narrow training windows, the GNN-SM framework provides a robust middle ground. It offers a more significant species reduction than DRGEP without the catastrophic loss of generalizability seen in the high-temperature specialized models.

\subsection{Iso-octane oxidation}\addvspace{10pt}

\begin{figure}[h!]
    \centering
    \includegraphics[width=180pt]{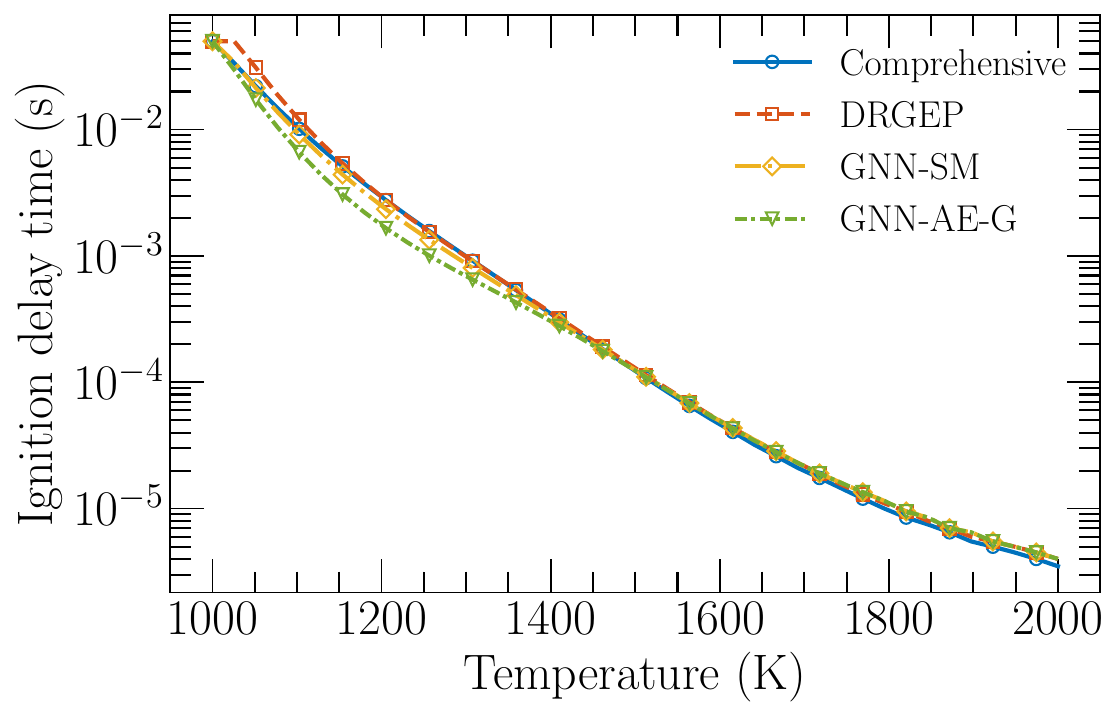}
    \caption{Validation of ignition delay times for reduced iso-octane mechanisms across conditions.}
    \label{fig:isooctane igt vs T}
\end{figure}

To further evaluate the proposed methodology on larger and more complex mechanisms, the iso-octane oxidation mechanism developed by Curran et al.~\cite{curran_comprehensive_2002} is considered, with training data obtained from the same thermochemical ranges used for methane and ethylene. This mechanism contains 1034 species and 8453 reactions and is widely used to describe higher hydrocarbon oxidation. A DRGEP reduction produces a mechanism with 63 species and 637 reactions, yielding an average IDT error of 8.81\% over the range $1000$--$2000$~K at atmospheric pressure ($1$~atm) and stoichiometric conditions ($\phi=1$). After seven reduction passes using GNN-SM, a mechanism containing 247 species and 2817 reactions is obtained. In contrast, the GNN-AE-G framework produces a significantly smaller mechanism with 55 species and 553 reactions, illustrating that maintaining accuracy over a broad thermochemical range requires the preservation of multiple reaction pathways. Targeting reduction around 1500~K enables a substantially more compact mechanism than DRGEP while maintaining excellent accuracy (IDT error $<5\%$) near the training region and at higher temperatures, with larger deviations at lower temperatures, as shown in Fig.~\ref{fig:isooctane igt vs T}.

\begin{table}[h]
\footnotesize
\centering
\caption{Comparison of mechanisms for iso-octane oxidation (1000K to 2000K).}
\label{curran mechanisms_comparison}
\begin{tabular}{lccc}
\hline
\textbf{Mechanism} & \textbf{$\mathbf{n_s/n_r}$} & \textbf{Avg. IDT Error (\%)} \\ \hline
Curran et al.   & 1034 / 8453  & --- \\
DRGEP      & 63 / 637  & 8.81 \\ \hline
\multicolumn{3}{l}{\textbf{GNN-SM}} \\ \hline
Pass 7        & 247 / 2817  & 9.19 \\ \hline
\multicolumn{3}{l}{\textbf{GNN-AE}} \\ \hline
GNN-AE-G    & 55 / 553   & 16.69 \\ \hline
\end{tabular}
\end{table}

\section{Conclusions\label{sec:conclusions}} \addvspace{10pt}

A graph neural network (GNN)-based framework for chemical mechanism reduction has been developed and validated on methane, ethylene, and iso-octane kinetic mechanisms. By combining graph-based representations of species-reaction networks with a pre-trained surrogate model, the GNN-SM approach identifies kinetically significant pathways while maintaining accuracy across a broad range of thermochemical states. This strategy enables mechanism size reductions of up to $70\%$ while preserving key ignition characteristics, demonstrating performance comparable to established graph-based reduction techniques. Alternatively, the GNN-AE framework employs an autoencoder identity-mapping strategy to learn compact representations of reaction networks within a restricted thermochemical regime, enabling more aggressive dimensionality reduction while maintaining fidelity within the trained state space. When coupled with greedy search refinement, the GNN-AE approach produces highly compact mechanisms, achieving reductions of up to $95\%$ in species and reactions, even for large mechanisms such as iso-octane. The dual implementation provides flexibility between robustness and compactness, enabling reductions tailored either for broad applicability or specific operating conditions. Beyond standalone mechanism reduction, the proposed framework can serve as an efficient automated precursor to expert-guided analytical reduction, significantly narrowing the search space and accelerating the development of reduced mechanisms for complex combustion systems.

\acknowledgement{CRediT authorship contribution statement} \addvspace{10pt}
\textbf{MNP}: conceptualization, methodology, software, formal analysis, investigation, writing (original draft).
\textbf{PD}: conceptualization, methodology, software, formal analysis, writing (original draft).
\textbf{KA}: conceptualization, methodology, formal analysis, supervision, writing (review and editing).

\acknowledgement{Declaration of competing interest} \addvspace{10pt}
The authors declare no competing financial interests or personal relationships.

\acknowledgement{Acknowledgments} 
\addvspace{10pt}
MNP acknowledges the support from RFPG scholarship. PD acknowledges the support from PMRF, Govt. of India. KA acknowledges funding support from the Core Research Grant (CRG/2023/007717) awarded by the Anusandhan National Research Foundation, India. The support provided by Aswin, Surya and Hridya in proofreading is duly acknowledged.

\footnotesize
\baselineskip 9pt

\clearpage
\thispagestyle{empty}
\bibliographystyle{journal}
\bibliography{main}

@article{curran_comprehensive_2002,
  author  = {H. J. Curran and P. Gaffuri and W. J. Pitz and C. K. Westbrook},
  title   = {A comprehensive modeling study of iso-octane oxidation},
  journal = {Combust. Flame},
  volume  = {129},
  number  = {3},
  year    = {2002},
  pages   = {253--280}
}

@article{narayanaswamy_component_2016,
  author  = {K. Narayanaswamy and H. Pitsch and P. Pepiot},
  title   = {A component library framework for deriving kinetic mechanisms for multi-component fuel surrogates: Application for jet fuel surrogates},
  journal = {Combust. Flame},
  volume  = {165},
  year    = {2016},
  pages   = {288--309}
}

@misc{smith_gri-mech_nodate,
  author = {G. P. Smith and D. M. Golden and M. Frenklach and N. W. Moriarty and B. Eiteneer and M. Goldenberg and C. T. Bowman and R. K. Hanson and S. Song and W. C. Gardiner and V. V. Lissianski and Z. Qin},
  title  = {{GRI}-{Mech} Home Page},
  note = {http://combustion.berkeley.edu/gri-mech/}
}

@article{lu_directed_2005,
  author  = {T. Lu and C. K. Law},
  title   = {A directed relation graph method for mechanism reduction},
  journal = {Proc. Combust. Inst.},
  volume  = {30},
  number  = {1},
  year    = {2005},
  pages   = {1333--1341}
}

@article{pepiot-desjardins_efficient_2008,
  author  = {P. Pepiot-Desjardins and H. Pitsch},
  title   = {An efficient error-propagation-based reduction method for large chemical kinetic mechanisms},
  journal = {Combust. Flame},
  volume  = {154},
  number  = {1},
  year    = {2008},
  pages   = {67--81}
}

@article{jaravel_error-controlled_2019,
  author  = {T. Jaravel and H. Wu and M. Ihme},
  title   = {Error-controlled kinetics reduction based on non-linear optimization and sensitivity analysis},
  journal = {Combust. Flame},
  volume  = {200},
  year    = {2019},
  pages   = {192--206}
}

@article{kipf_semi-supervised_2017,
  author  = {T. N. Kipf and M. Welling},
  title   = {Semi-Supervised Classification with Graph Convolutional Networks},
  journal = {arXiv preprint arXiv:1609.02907},
  year    = {2017}
}

@article{velickovic_graph_2018,
  author  = {P. Veli\v{c}kovi\'{c} and G. Cucurull and A. Casanova and A. Romero and P. Li\`{o} and Y. Bengio},
  title   = {Graph Attention Networks},
  journal = {arXiv preprint arXiv:1710.10903},
  year    = {2018}
}

@article{shi_masked_2021,
  author  = {Y. Shi and Z. Huang and S. Feng and H. Zhong and W. Wang and Y. Sun},
  title   = {Masked Label Prediction: Unified Message Passing Model for Semi-Supervised Classification},
  journal = {arXiv preprint arXiv:2009.03509},
  year    = {2021}
}

@article{sen_large_2010,
  author  = {B. A. Sen and E. R. Hawkes and S. Menon},
  title   = {Large eddy simulation of extinction and reignition with artificial neural networks based chemical kinetics},
  journal = {Combust. Flame},
  volume  = {157},
  number  = {3},
  year    = {2010},
  pages   = {566--578}
}

@article{borde_mathematical_2025,
  author  = {H. S{\'a}ez de Oc{\'a}riz Borde and M. Bronstein},
  title   = {Mathematical Foundations of Geometric Deep Learning},
  journal = {arXiv preprint arXiv:2508.02723},
  year    = {2025}
}

@article{sun_path_2010,
  author  = {W. Sun and Z. Chen and X. Gou and Y. Ju},
  title   = {A path flux analysis method for the reduction of detailed chemical kinetic mechanisms},
  journal = {Combust. Flame},
  volume  = {157},
  number  = {7},
  year    = {2010},
  pages   = {1298--1307}
}

@article{scarselli_graph_2009,
  author  = {F. Scarselli and M. Gori and A. C. Tsoi and M. Hagenbuchner and G. Monfardini},
  title   = {The graph neural network model},
  journal = {IEEE Trans. Neural Netw.},
  volume  = {20},
  number  = {1},
  year    = {2009},
  pages   = {61--80}
}

@article{rieth_direct_2024,
  author  = {M. Rieth and A. Gruber and E. R. Hawkes and J. H. Chen},
  title   = {Direct numerical simulation of low-emission ammonia rich-quench-lean combustion},
  journal = {Proc. Combust. Inst.},
  volume  = {40},
  number  = {1},
  year    = {2024},
  pages   = {105558}
}

@article{niemietz_direct_2023,
  author  = {K. Niemietz and L. Berger and M. Huth and A. Attili and H. Pitsch},
  title   = {Direct numerical simulation of flame-wall interaction at gas turbine relevant conditions},
  journal = {Proc. Combust. Inst.},
  volume  = {39},
  number  = {2},
  year    = {2023},
  pages   = {2209--2218}
}

@inproceedings{dash_analysis_nodate,
  author    = {P. Dash and K. Aditya and H. Kolla and J. H. Chen},
  title     = {An analysis of lean premixed ethylene-air flame stabilization over a backward facing step},
  booktitle = {AIAA AVIATION 2023 Forum},
  year      = {2023}
}

@misc{ZDV2023,
  author = {Y. Zhang and W. Dong and L. Vandewalle and R. Xu and G. Smith and H. Wang},
  title  = {Foundational Fuel Chemistry Model Version 2.0 (FFCM-2)},
  year   = {2023},
  note   = {https://web.stanford.edu/group/haiwanglab/FFCM2}
}

@article{dash_super-resolution_2026,
  author  = {P. Dash and K. Aditya and C. E. Frouzakis and M. Bode},
  title   = {Super-resolution of turbulent reacting flows on complex meshes using graph neural networks},
  journal = {arXiv preprint arXiv:2603.01080},
  year    = {2026}
}

@article{li_graph_2022,
  author  = {Z. Li and K. Meidani and P. Yadav and A. Barati Farimani},
  title   = {Graph neural networks accelerated molecular dynamics},
  journal = {J. Chem. Phys.},
  volume  = {156},
  number  = {14},
  year    = {2022},
  pages   = {144103}
}

@article{kipf2016variationalgraphautoencoders,
  author  = {T. N. Kipf and M. Welling},
  title   = {Variational Graph Auto-Encoders},
  journal = {arXiv preprint arXiv:1611.07308},
  year    = {2016}
}


\newpage

\small
\baselineskip 10pt


\end{document}